\pdfoutput=1
\documentclass[11pt]{article}
\usepackage{amsmath}
\usepackage[final]{acl}

\usepackage{times}    
\usepackage{latexsym}
\usepackage{booktabs}
\usepackage{tabularx}
\usepackage{float}
\usepackage{enumitem}
\usepackage{caption}
\usepackage{graphicx}
\usepackage{dblfloatfix}
\usepackage[T1]{fontenc}
\usepackage[utf8]{inputenc}

\usepackage{microtype}

\usepackage{inconsolata}

\usepackage{graphicx}
\setlist[itemize]{noitemsep, topsep=0pt, partopsep=0pt, parsep=0pt}
\title{Hybrid Fact-Checking that Integrates Knowledge Graphs, Large Language \\Models, and Search-Based Retrieval Agents Improves Interpretable Claim Verification}

\author{
Shaghayegh Kolli,\thanks{These authors contributed equally.} \
Richard Rosenbaum,\footnotemark[1] \
Timo Cavelius, \\
\textbf{Lasse Strothe,} \
\textbf{Andrii Lata,} \
\textbf{Jana Diesner} \\
Technical University Munich \\
\small
{(shaghayegh.kolli, richard.rosenbaum, timo.cavelius, lasse.strothe, andrii.lata, jana.diesner)@tum.de}
}

\begin{document}
\maketitle
\begin{abstract}

Large language models (LLMs) excel in generating fluent utterances but can lack reliable grounding in verified information. At the same time, knowledge-graph-based fact-checkers deliver precise and interpretable evidence, yet suffer from limited coverage or latency. By integrating LLMs with knowledge graphs and real-time search agents, we introduce a hybrid fact-checking approach that leverages the individual strengths of each component. Our system comprises three autonomous steps: 1) a Knowledge Graph (KG) Retrieval for rapid one‑hop lookups in DBpedia, 2) an LM-based classification guided by a task-specific labeling prompt, producing outputs with internal rule-based logic, and 3) a Web Search Agent invoked only when KG coverage is insufficient. Our pipeline achieves an F1 score of 0.93 on the FEVER benchmark on the Supported/Refuted split without task‑specific fine‑tuning. To address \textit{Not enough information} cases, we conduct a targeted reannotation study showing that our approach frequently uncovers valid evidence for claims originally labeled as \textit{Not Enough Information (NEI)}, as confirmed by both expert annotators and LLM reviewers. With this paper, we present a modular, open-source fact-checking pipeline with fallback strategies and generalization across datasets. 

\end{abstract}

\section{Introduction}
\begin{figure*}[!b]
  \centering
  \includegraphics[width=\textwidth]{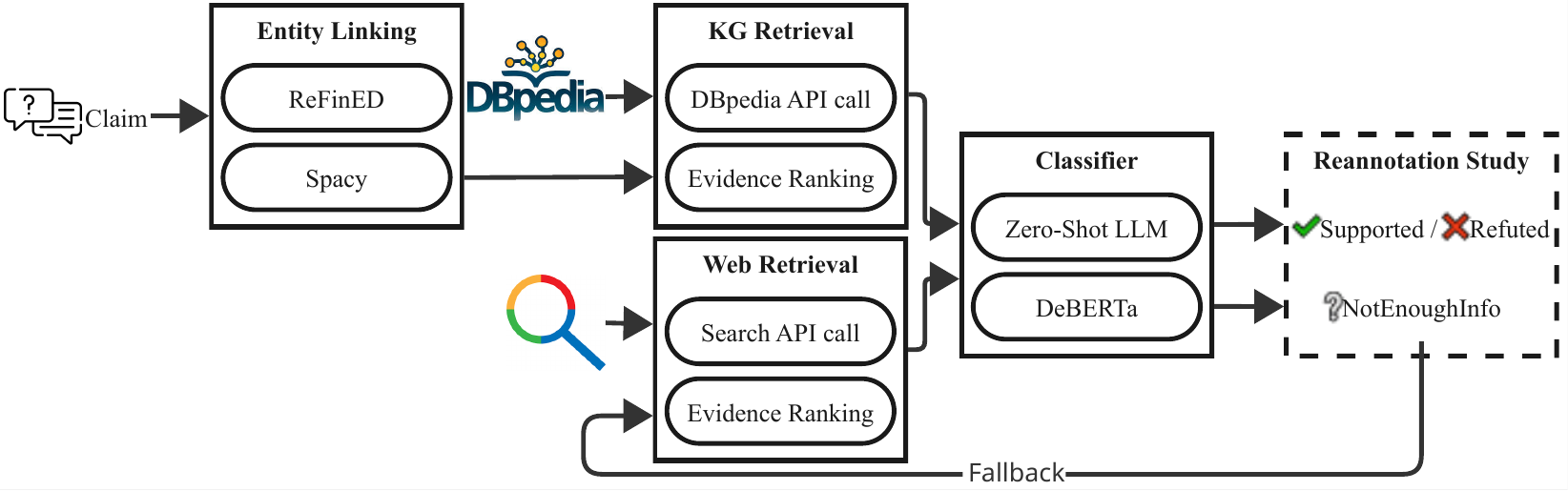}
  \caption{Hybrid fact‑verification pipeline: a KG‑first pass links entities
  to Wikidata Q‑IDs, retrieves and ranks one‑hop DBpedia triples for
  classification; NEI outputs trigger a Web‑RAG fallback that rewrites the
  claim, retrieves web snippets, and re‑evaluates with the same model. Ambiguous NEI cases are validated by human annotators.}
  \label{fig:pipeline}
\end{figure*}
LLMs have advanced knowledge-intensive NLP tasks, but can generate ungrounded or hallucinated content, which undermines their reliability for automated fact checking \cite{brown2020language}. Knowledge-graph (KG)-based systems can provide explicit and transparent evidence through structured triples, but remain restricted due to their limited coverage and slower response times in open‑domain scenarios \cite{jiang-etal-2020-hover,kim-etal-2023-factkg}.
Recent work, such as Generate-on-Graph \cite{xu-etal-2024-generate}, treats LLMs as agents that generate missing KG triples, highlighting the potential of hybrid agent–KG reasoning frameworks.

This paper asks how a modular hybrid system can make fact-checking more reliable, and shows how a real-time pipeline improves both coverage and interpretability.
We propose a real-time, agent-based pipeline (Figure~\ref{fig:pipeline}) that integrates three autonomous steps: 1) a KG Retrieval for rapid one‑hop lookups in DBpedia \cite{lehmann2015dbpedia}; 2) Language models to classify claims with a task-specific classification prompt using labels such as \textit{Supported}, \textit{Refuted}, or \textit{Not Enough Information} \textit{(NEI)} \cite{wei2022chain}; and 3) a Web Search Agent invoked only when \textit{NEI} is returned, rewriting the claim for on-demand retrieval \cite{lewis2020rag,tan2023evidence}.\footnote{The implementation is open source and on GitHub at  
\href{https://github.com/AndriiLata/aiFactCheck}{github.com/AndriiLata/aiFactCheck}.} While our system does not perform multi-hop reasoning, it remains modular across evidence types (structured KG evidence, unstructured web evidence), using retrieval to compensate for KG’s single-hop limitations. This KG‑first, web‑adaptive strategy leverages the explainability of structured data while preserving open‑domain coverage.

We evaluated our approach on the FEVER benchmark \cite{thorne-etal-2018-fever}, its adversarial extension FEVER 2.0 \cite{thorne-etal-2019-fever2}, and, that is, the FactKG dataset \cite{kim-etal-2023-factkg}, achieving up to 0.93 F1 on FEVER and competitive results across all three without task-specific tuning. A focused \textit{Not Enough Information} reannotation study shows that our pipeline can uncover valid evidence for claims labeled as unverifiable, a finding corroborated by both expert human annotators and LLM reviewers.

\section{Related Work}
Recent work in automated fact verification has focused on integrating structured knowledge sources, retrieval components, and LLMs to improve factual consistency and evidence grounding ~\cite{cao2025enhancing, opsahl-2024-fact, kim-etal-2023-kg}. A growing number of systems have been combining neural models with KGs~\cite{zhou-etal-2019-gear, kim-etal-2023-factkg, yao2019kgbert} or using web-based retrieval to expand coverage~\cite{chen-etal-2024-complex}.

KG-based methods often rely on symbolic triples of the form \texttt{(subject, predicate, object)} as evidence. Prior studies have explored how to align natural language claims with KG facts using embedding models~\cite{yao2019kgbert}, graph-based reasoning~\cite{zhou-etal-2019-gear}, semantic matching between claims and triples~\cite{kim-etal-2023-factkg}, and LLMs~\cite {kim-etal-2023-kg}. While KGs offer structured and interpretable evidence, they can be limited by coverage and connectivity, particularly for claims requiring multi-hop or commonsense reasoning ~\cite{peng2023knowledge}.

In contrast, web-based fact-checking systems retrieve textual evidence from open-domain sources. OE-Fact~\cite{tan2023evidence}, for instance, used LLMs to process retrieved snippets and generate decisions. Retrieval-augmented generation (RAG)~\cite{lewis2020rag} has also been applied to fact verification tasks by conditioning generation on retrieved content. However, reliance on web-based, unstructured evidence raises concerns around evidence quality and verifiability.

There is a growing interest in agent-based and modular architectures for fact verification. The FIRE system~\cite{xie-etal-2025-fire} employs an iterative retrieval and verification process, where the model dynamically decides whether to retrieve more evidence or make a decision. Such approaches reflect a broader trend toward separating evidence retrieval from claim evaluation, often across different evidence sources or reasoning stages ~\cite{zhang-etal-2023-relevance}.
Finally, several studies have pointed out limitations with benchmark labels, particularly in the NEI category ~\cite{hu2024towards}. Prior work has shown that some NEI claims can be verified with external evidence~\cite{schuster-etal-2019-towards}, highlighting the role of human judgment in evaluating evidence sufficiency and the need for annotation guidelines that reflect real-world complexity.

To expand on this prior work, we developed a modular pipeline that combines structured KG evidence with an agent fallback retrieval and includes an interpretable classification component. 

\section{Methodology}
Given a natural language claim $C$, our goal is to predict a label $Y \in \{\textsc{Supported}, \textsc{Refuted}, \textsc{NEI}\}$, along with a small set of textual or structured evidence $E^*$ that justifies the decision. Our system follows a two-stage architecture: a KG-first classification stage, followed by a fallback retrieval and reasoning stage using open-domain web evidence. The system does not require task-specific training and operates in a zero-shot inference mode.
An overview of the pipeline is shown in Figure~\ref{fig:pipeline}.

\noindent\textbf{Stage 1: Knowledge Graph First Pass}

\textit{Entity linking:}
We use ReFinED~\citep{ayoola-etal-2022-refined} to detect and disambiguate named-entity mentions in the claim $c$, mapping each surface span to a Wikidata Q-ID~\citep{vrandevcic2014wikidata}; if none is produced, we fall back to spaCy’s EntityLinker~\citep{honnibal2020spacy}. Resolved IDs are mapped to DBpedia via \texttt{owl:sameAs}~\citep{dbpedia}. Many synonyms and paraphrases are covered through surface-form dictionaries via ReFinED and Wikipedia redirects, but it does not handle arbitrary paraphrases. In case no Wikidata ID can be assigned, the mention is skipped in the KG stage but may still be handled by the fallback.

\textit{Triple retrieval:}
For each linked entity $e$, we issue a one-hop SPARQL~\citep{prud2006sparql} query to extract all RDF triples $t = \langle s, p, o \rangle \in \text{DBpedia}$ where $s = e$ or $o = e$. For example, one triple could look like this: "Barack\_Obama -> birthPlace -> Hawaii". We exclude triples with metainformation predicates using a handcrafted blacklist.

\textit{Triple scoring:}
Each candidate triple $t$ is paired with the original claim and scored for semantic relevance using the \texttt{ms-marco-MiniLM-L6-v2} cross-encoder~\citep{wang2020minilm}. The input format for this is $[\text{CLS}]\, C\, [\text{SEP}]\, t\, [\text{SEP}]$. We retain the top $k=5$ highest-scoring triples, denoted as $E^*_{\text{KG}} = \{t_1, \dots, t_k\}$.

\textit{KG classification:}
The set $\{C\} \cup E^*_{\text{KG}}$ is passed to either a GPT‑4o mini~\citep{openai2024gpt4o_mini} instance or a DeBERTa‑v3 MNLI \cite{hedebertav3} model instance. The model assigns a local label $y_{\text{KG}} \in \{\textsc{S}, \textsc{R}, \textsc{N}\}$ and provides a justification based on the supporting evidence triples. If $y_{\text{KG}} \in \{\textsc{S}, \textsc{R}\}$, the pipeline terminates and outputs $Y = y_{\text{KG}}$. Otherwise, we proceed to Stage 2.

\noindent\textbf{Stage 2: Web-Based Fallback}

\textit{Query rewriting:}
For cases labeled \textsc{Not Enough Info}, we prompt GPT-4o mini to paraphrase the original claim into 3--5 high-recall search queries. These are submitted to the Google Programmable Search API \cite{google2025customsearch}.

\textit{Snippet retrieval:}
The top $n \leq 100$ web snippets are collected. Each snippet $s_j$ is scored with the same MiniLM cross-encoder as in Stage 1. We retain the top $k=5$ snippets, forming $E^*_{\text{Web}} = \{s_1, \dots, s_k\}$.

\textit{Evidence classification:}
Each $(C, s_j)$ pair is classified using a modular verifier—either a zero‑shot LLM (GPT‑4o mini) or a DeBERTa‑v3 MNLI model—with all configuration details deferred to Section~\ref{sec:impl}.
The final verdict is $Y = y_{\text{Web}}$ and $y_{\text{Web}} \in \{\textsc{Supported}, \textsc{Refuted}, \textsc{NEI}\}$. If \textsc{NEI} is returned as the output, the fallback mechanism is not triggered again. When the pipeline was configured with an LLM and DeBERTa, we observed that the fallback mechanism was invoked in about 23\% of all test cases.

% \subsection{Final Decision Logic}
% Let $y_{\text{KG}}$ and $y_{\text{Web}}$ be the local labels from the KG and web stages, respectively. We define the final verdict $Y$ by the following rule:

% \begin{itemize}[noitemsep]
%     \item If $y_{\text{KG}} \in \{\textsc{Supported}, \textsc{Refuted}\}$, return $Y = y_{\text{KG}}$ and $E^*_{\text{KG}}$.
%     \item Else if $y_{\text{Web}} \in \{\textsc{Supported}, \textsc{Refuted}\}$, return $Y = y_{\text{Web}}$ and $E^*_{\text{Web}}$.
%     \item Else, return $Y = \textsc{Not Enough Information}$.
% \end{itemize}
% This logic prioritizes interpretable structured evidence, but falls back to broader open-domain information when the KG lacks coverage.

\section{Implementation}
\label{sec:impl}
Our system is built in a modular way so that it can be accessed through a simple REST interface \cite{fielding2000architectural}. 
The modularity makes it easy to test different components or replace models. We experiment with two evidence classifiers:

\textbf{GPT‑4o mini (LLM):}
For each evidence item $e_i$, we construct a JSON prompt containing the claim $c$ and the list $\{e_i\}$ (triples or snippets).  
The model returns
\(\{\text{``label'': }\mathsf{S|R|N},\,\text{``reason'': }r\}\),
where \(r\) is a single sentence that cites evidence. During development, we tested various LLM prompt variants to maximize classification accuracy and robustness before settling on the final versions reported in our results. The final prompts can be found in the appendix ~\ref{sec:appendixPr}.

\textbf{DeBERTa‑v3‑MNLI:}
We cast fact verification as natural‑language inference.  
Every pair \(\langle c,e_i\rangle\) is transformed into
\([\text{CLS}]\, e_i\, [\text{SEP}]\, c\, [\text{SEP}]\).  
The model \cite{hedebertav3} outputs logits $(\ell_E,\ell_N,\ell_C)$ for \{\textsc{Entailment}, \textsc{Neutral}, \textsc{Contradiction}\}.  
We apply \texttt{softmax} and pick the label with the highest probability $p_{\max}$. Afterwards, we map them back to the FEVER labels.

\textbf{Datasets:} For our main experiments, we use the FEVER dataset, which labels claims as Supported, Refuted, or Not Enough Information. To ensure fair comparison across experiments and with other papers and avoid ambiguity, we randomly sample 1,000 FEVER claims, explicitly removing all NEI-labeled instances.

\section{Results and Discussion}

Table~\ref{tab:fact_checking_results} reports the standard NLP accuracy evaluation metrics of precision, recall, and F$_1$ across (i) claim-only baselines, (ii) single source stages (KG only or Web only), and (iii) the complete two-stage pipeline. Three annotated output examples are provided in Appendix~\ref{sec:appendixEx}.

\textbf{Baselines: }
Following the claim-only setting in prior work, zero-shot LLMs without retrieval can resolve a portion of FEVER claims but remain ungrounded. The best baseline here (Zero-Shot 4o-mini) results in an F$_1$~0.801, while Zero-Shot 4.1-nano leads to F$_1$~0.734. Although these models are competitive, the absence of explicit evidence limits the verifiability of their reasoning.

\textbf{Separate Stages:}
Single-source variants show opposing error profiles. KG-only with an LLM results in high precision (0.944) but lower recall (0.734), reflecting reliable yet sparse coverage. 

In contrast, web-only configurations are more balanced (e.g., LLM Web-only: Prec.~0.912, Rec.~0.908), suggesting broader coverage at the cost of increased noise.
\begin{table}[htbp]
\centering
\setlength{\tabcolsep}{3pt}
\renewcommand{\arraystretch}{1.1}
\begin{tabular}{@{}lccc@{}}
    \toprule
    \textbf{Model Variant} & \textbf{Prec.} & \textbf{Rec.} & \textbf{F1} \\
    \midrule
    \multicolumn{4}{l}{\textit{Baselines}} \\
    Random Choice          & 0.500 & 0.500 & 0.500 \\
    BERT-Base (no ret.)    & 0.649     & 0.594     & 0.620 \\
    Zero Shot 4.1 nano\footnotemark[1]     & 0.816 & 0.720 & 0.734 \\
    Zero Shot 4o mini\footnotemark[2]      & \textbf{0.826} & \textbf{0.790} & \textbf{0.801} \\
    \midrule
    \multicolumn{4}{l}{\textit{Separate Stages}} \\
    KG alone, LLM          & \textbf{0.944} & 0.734 & 0.826 \\
    KG alone, DEBERTA      & 0.882 & 0.620 & 0.714 \\
    Web only, LLM          & 0.912 & \textbf{0.908} & \textbf{0.909} \\
    Web only, DEBERTA      & 0.913 & 0.878 & 0.895 \\
    \midrule
    \multicolumn{4}{l}{\textit{Full Pipeline}} \\
    LLM, LLM               & 0.920 & 0.916 & 0.917 \\
    DEBERTA, LLM           & 0.883 & 0.853 & 0.859 \\
    LLM, DEBERTA           & \textbf{0.930} & \textbf{0.926} & \textbf{0.927} \\
    DEBERTA, DEBERTA       & 0.887 & 0.849 & 0.860 \\
    \midrule
    \multicolumn{4}{l}{\textit{Stronger LLM 4.1 Mini\footnotemark[1]}} \\
    LLM, LLM               & \textbf{0.932} & \textbf{0.931} & \textbf{0.931} \\
    LLM, DEBERTA           & 0.919 & 0.899 & 0.908 \\
    \bottomrule
\end{tabular}
\caption{Performance comparison of model variants on FEVER. \textsuperscript{1}\cite{openai2025gpt4_1}, \textsuperscript{2}\cite{openai2024gpt4o_mini}}
\label{tab:fact_checking_results}
\end{table}\\
\textbf{Full pipeline:}
Combining KG-first inference with a web fallback led to the highest overall performance among the configurations evaluated. Using the baseline language model (GPT-4o-mini), the full pipeline incorporating a downstream DEBERTA classifier resulted in an F$_1$ score of approximately 0.927, compared to 0.917 with the language model alone. Substituting the language model with GPT-4.1-mini further increases the F$_1$ score to 0.931. Consistent with prior work \cite{li-etal-2024-self}, our pipeline maintains stable performance across different classifier configurations and benefits from increased model capacity.

\textbf{Design Choice:}
We adopt a KG-first approach to prioritize precision and interpretability, resorting to Web retrieval only when KG evidence is insufficient (NEI). This design choice improves transparency by grounding decisions in structured evidence and reducing unnecessary web queries.
\begin{table}[H]
\centering
\setlength{\tabcolsep}{3pt}
\renewcommand{\arraystretch}{1.1}
\begin{tabular}{@{}p{3.5cm}ccc@{}}
    \toprule
    \textbf{Dataset} & \textbf{Prec.} & \textbf{Rec.} & \textbf{F1} \\
    \midrule
    FEVER 2.0          & 0.797 & 0.769 & 0.783 \\
    FactKG            & 0.791     & 0.757     & 0.774 \\
    \bottomrule
\end{tabular}
\caption{Performance on other fact-checking datasets.} 
\label{tab:fact_checking_other_datasets}
\end{table}
\textbf{Comparisons:} Without task‑specific fine‑tuning, our pipeline transfers well to FEVER~2.0 (F$_1$=0.78) and FactKG (F$_1$=0.77). These results can be seen in table \ref{tab:fact_checking_other_datasets}. 
\begin{table}[H]
\centering
\setlength{\tabcolsep}{3pt}
\renewcommand{\arraystretch}{1.1}
\begin{tabular}{@{}p{4.5cm}cc@{}}
    \toprule
    \textbf{Results} & \textbf{Mode} & \textbf{Acc.} \\
    \midrule
    FEVER, Ours  & S/R & 0.931 \\
    \cite{lewis2020rag}            & S/R     & 0.895\\
    \midrule
    FEVER, Ours  & S/R/N & 0.702 \\
    \cite{tan2023evidence}            & S/R/N     & 0.542\\
    \midrule
    FEVER 2.0, Ours  & S/R & 0.732 \\
    \cite{yuan-vlachos-2024-zero}            & S/R     & 0.733\\
    \bottomrule
\end{tabular}
\caption{Direct comparisons to other related work.} 
\label{tab:fact_checking_comparisons}
\end{table}
In the context of recent systems using open‑domain retrieval and LLMs, prior work reports 89.5\% S/R on FEVER with Wikipedia retrieval and a seq2seq verifier~\cite{lewis2020rag}; \citeauthor{yuan-vlachos-2024-zero} reported 73.34\% S/R on FEVER~2.0 via zero‑shot triple extraction and KG retrieval, which we match (73\%); and \citeauthor{tan2023evidence} reported 54.2\% S/R/N on FEVER with web evidence, which we exceed even without considering NEI (results in table \ref{tab:fact_checking_comparisons}.

\subsection{Analysis of NEI-Labeled Claims}
A recurring issue in FEVER involves NEI labels for which our system nonetheless retrieves supporting or refuting evidence. To further examine this, we constructed a targeted evaluation: we randomly sampled 150 NEI claims where our model consistently surfaced evidence and asked two human annotators and one LLM to judge evidence sufficiency (Appendix~\ref{sec:appendixNEI}).

Over 70\% of cases were deemed \emph{sufficient} by at least one human, indicating that the pipeline retrieves meaningful evidence for many claims labeled NEI. Inter-annotator agreement was moderate: Fleiss' $\kappa$ among humans was 0.385 (compare Figure \ref{fig:agreement_score} in Appendix \ref{sec:appendixNEI}), with unanimous agreement in 70.7\% of instances; LLM-human agreement varied (compare Figure \ref{fig:agreement_score}, reflecting the subjectivity of sufficiency judgments. These findings suggest that assessing sufficiency depends on annotator strictness and perceived completeness of the evidence. Including more annotators, reconciliation among human annotators,  and a broader range of NEI cases could strengthen the reliability of these conclusions. Despite variability, the $>$70\% sufficiency rate (cf. Fig.~\ref{fig:suff_rate} in Appendix~\ref{sec:appendixNEI}) suggests that our pipeline reliably finds relevant evidence. Thus, excluding NEI from baseline comparisons is methodologically justified under our setup.

\section{Conclusion and Future Work}
We present a real-time fact-checking pipeline that combines the strengths of KGs and web retrieval to address the limitations of existing LLM-based and KG-based systems. Our KG-first, web-adaptive approach delivers both high precision and broad coverage, achieving strong empirical results across FEVER and other standard benchmarks without task-specific fine-tuning. It offers competitive accuracy with stronger reliability and interpretability than purely web‑based or neural setups. In addition, our NEI re-annotation study shows that in over 70\% of cases, the system retrieves meaningful evidence for claims originally labeled \textit{Not Enough Information}. However, subjectivity in human judgments remains a challenge.

Overall, our work demonstrates the value of integrating structured and unstructured evidence for robust, interpretable open-domain fact verification. For future work, we plan to enhance support for multi-hop evidence, improve the detection of truly unverifiable claims, explore alternative classifiers, and extend our approach to additional knowledge sources and datasets.

\section*{Limitations}
While our KG-first, web-adaptive pipeline achieves strong performance and generalizes well across benchmarks, several limitations remain.

Retrieving multi-hop evidence from KGs is still a major challenge. Our system mainly uses single-hop paths for speed and coverage, but more complex claims may require combining information from multiple nodes or documents, which is not fully captured by our current approach.

The pipeline is also sensitive to error propagation from early components into the pipeline; a long-standing issue in pipelines from NLP tasks to downstream applications \cite{diesner2015impact}. Small mistakes in entity linking, predicate selection, or evidence ranking can propagate through the system and lead to incorrect final labels. This suggests that improving component accuracy, especially early on in the upstream parts, could further enhance overall system reliability.

Additionally, our method assumes that either supporting or refuting evidence can always be found in the KG or on the web. As a result, the system currently has no mechanism for properly handling NEI claims and cannot explicitly indicate when evidence is missing. This limits its applicability to datasets where NEI is a significant or required label.

Finally, by emphasizing broad coverage and adaptability for open-domain fact-checking, the system trades off a few SOTA points on specific, specialized benchmarks. This reflects design choices made to favor practical, real-time usage over narrow optimization.

\section*{Ethical Considerations}
Developing automated fact-checking systems involves several ethical challenges, particularly around fairness, transparency, and reliability. Our pipeline relies on data from public KGs and accessible (in the sense of visible) web sources, which may contain biases, errors, misinformation, and a lack of diverse perspectives, and relies on the provision of these data by others, which may imply intellectual property constraints that limit their use depending on jurisdiction and use case. These limitations can influence both evidence retrieval and final predictions. Users are responsible for copyright compliance, and we recommend favoring open-access sources.
A key part of our evaluation involved human annotation. We recruited two graduate students with strong English proficiency and familiarity with research ethics. Annotators participated in structured training sessions to ensure consistent application of our guidelines. Their judgments in the NEI reannotation study highlighted the subjectivity involved in assessing evidence sufficiency and underscored the importance of incorporating human input when evaluating model outputs.
Our system currently does not explicitly model uncertainty or signal when evidence is insufficient, which can lead to overconfident predictions in cases beyond the scope of available sources. Additionally, biases in benchmark datasets, including claim selection and annotation practices, can impact generalizability. 

\bibliography{latex/anthology, latex/custom}

\newpage
\appendix

\section{Labeled Output Examples}
\label{sec:appendixEx}

\textbf{Example 1:}
\vspace{0.3em}\\
Claim: “Eric Trump's father is banned from ever becoming president.”\\
True Label: \textbf{Refuted}\\
Entities: Eric\_Trump, President, Father\\
System: NEI $\rightarrow$ Web Search $\rightarrow$ \textbf{Refuted}\\
Explanation: Snippet 2 indicates Donald Trump is a President-Elect, so he is eligible to become president.\\
Evidence: "Eric Trump, the second son of President-Elect Donald Trump, told The Post this week his father has a long to-do list ready for his White" \cite{durbin2024nypost-eric-trump}

\vspace{1em}

\noindent\textbf{Example 2:}
\vspace{0.3em}\\
Claim: “Black Mirror is a British science fiction television series about modern society.”\\
True Label: \textbf{Supported}\\
Entities: Black\_Mirror, Television\_in\_the\_United\_Kingdom, Science\_Fiction\\
System: \textbf{Supported}\\
Explanation: Path 1 confirms Black Mirror is a British anthology television series exploring science fiction themes about modern society.\\
Evidence: Path 1: Black\_Mirror $\rightarrow$ Abstract

\vspace{1em}

\noindent\textbf{Example 3:}
\vspace{0.3em}\\
Claim: “Arya Stark was created by George R. R. Martin.”\\
True Label: \textbf{Supported}\\
Entities: Arya\_Stark, George\_R.\_R.\_Martin\\
System: \textbf{Supported}\\
Explanation:  Path 1 directly records creator George R. R. Martin for Arya Stark.\\
Evidence: Path 1: Arya\_Stark $\rightarrow$ creator $\rightarrow$ George\_R.\_R.\_Martin

\vspace{1em}

\section{Classifier prompts}
\label{sec:appendixPr}

\subsection*{LLM prompt for KG stage}
\small

\noindent\textbf{System Prompt (static)}
\begin{flushleft}
\ttfamily
You are a world-class fact-verification assistant.\\

Given a claim and a numbered list of evidence paths, choose exactly one label:\\
\quad • Supported – at least one path exactly affirms \\
\quad\quad the claim’s assertion.\\
\quad • Refuted – at least one path \textit{explicitly} \\
\quad\quad contradicts it (e.g. predicate like “is \\
\quad\quad not”).\\
\quad • Not Enough Info – otherwise.\\

Rules:\\
1. If any path affirms the claim’s predicate+object, label Supported.\\
2. Only label Refuted if a path uses negation or clear contradiction.\\
3. Otherwise label Not Enough Info.\\
4. Use only the provided paths; do NOT invent facts.\\
5. Keep reasoning private — do NOT show chain-of-thought.\\
6. Output only a single JSON object:\\
\{\\
\ \ \ "label": <Supported\textbar Refuted\textbar Not Enough Info>,\\
\ \ \ "reason": <one concise sentence citing path number(s)>\\
\}
\end{flushleft}

\vspace{1ex}
\noindent\textbf{User Prompt (input)}
\begin{flushleft}
\ttfamily
Claim: <CLAIM>\\

Evidence paths:\\
<EVIDENCE\_PATHS>\\

Instruction:\\
- Label Supported if any path’s predicate and object exactly match the claim.\\
- Label Refuted only if a path explicitly contradicts (uses “not”, “no”, etc.).\\
- Otherwise label Not Enough Info.\\

Examples:\\

1) Supported\\
Claim: “Alice’s birthplace is Canada.”\\
1. Alice → birthPlace → Canada\\
Output:\\
\{"label":"Supported", "reason":"Path 1 exactly matches birthPlace→Canada."\}\\

2) Refuted\\
Claim: “Bob is an exponent of Doom metal.”\\
1. Bob → is not an exponent of → Doom\_metal\\
Output:\\
\{"label":"Refuted", "reason":"Path 1 explicitly states ‘is not an exponent of Doom metal’."\}\\

3) Not Enough Info\\
Claim: “Carol’s nationality is Spanish.”\\
1. Carol → birthPlace → Barcelona\\
Output:\\
\{"label":"Not Enough Info", "reason":"Path 1 does not confirm nationality."\}
\end{flushleft}

\normalsize

\subsection*{LLM prompt for Web-Search stage}
\small

\noindent\textbf{System Prompt (static)}
\begin{flushleft}
\ttfamily
You are a world-class fact-verification assistant.\\

Your job: given a claim and a small numbered list of evidence snippets, decide only one of two labels:\\
\quad • Supported – at least one snippet clearly\\ 
\quad\quad confirms the claim.\\
\quad • Refuted – at least one snippet explicitly\\
\quad\quad contradicts the claim.\\

You must not output any other label.\\
Use only the provided snippets; do not invent facts or fetch external data.\\ Keep your reasoning private — do not expose chain-of-thought.\\

Output exactly one JSON object:\\
\{\\
\ \ \ "label": <Supported\textbar Refuted>,\\
\ \ \ "reason": <one short sentence citing snippet number(s)>\\
\}
\end{flushleft}

\vspace{1ex}
\noindent\textbf{User Prompt (input)}
\begin{flushleft}
\ttfamily
Claim: <CLAIM>\\

Evidence snippets:\\
<EVIDENCE\_SNIPPETS>\\

Instruction:\\
- If any snippet affirms the claim’s exact assertion, label Supported.\\
- If any snippet contradicts it (negation, opposite fact), label Refuted.\\
- You must choose one of the two — no other options.\\

Examples:\\

Supported Example:\\
Claim: “Alice’s birthplace is Canada.”\\
1. Alice → birthPlace → Canada\\
Output:\\
\{"label":"Supported", "reason":"Snippet 1 shows birthPlace → Canada."\}\\

Refuted Example:\\
Claim: “Bob is an exponent of Doom metal.”\\
1. Bob → is not an exponent of → Doom metal\\
Output:\\
\{"label":"Refuted", "reason":"Snippet 1 states 'is not an exponent of Doom metal'."\}
\end{flushleft}

\normalsize

\subsection*{LLM prompt for zero-shot baselines}
\small

\noindent\textbf{System Prompt (static)}
\begin{flushleft}
\ttfamily
You are a world-class fact checker. You will receive a claim, and your job is to verify its factual accuracy based only on your knowledge.\\

You must choose one of two labels:\\
\quad • Supported – the claim is clearly true.\\
\quad • Refuted – the claim is clearly false.\\

If unsure, make your best guess. Avoid using vague language.\\

Output exactly one JSON object like this:\\
\{\\
\ \ \ "label": "Supported" or "Refuted",\\
\ \ \ "reason": "short explanation of why you chose this label"\\
\}
\end{flushleft}

\vspace{1ex}
\noindent\textbf{User Prompt (input)}
\begin{flushleft}
\ttfamily
Claim: <CLAIM>\\

Decide whether this is Supported or Refuted.
\end{flushleft}

\normalsize

\subsection*{Prompt for Web-Search Paraphrasing}
\small

\noindent\textbf{System Prompt (static)}
\begin{flushleft}
\ttfamily
You are an expert fact-checking assistant who writes superb web-search queries.\\
Given a claim, reformulate it into 3–5 concise, high-recall search queries. Each query should:\\
\quad • be under 12 words\\
\quad • keep critical named entities, dates, and \\
\quad\quad numbers\\
\quad • add quotation marks for exact phrases when \\
\quad\quad helpful\\
\quad • avoid hashtags or advanced operators other \\
\quad\quad than quotes\\

Return exactly one JSON object like this:\\
\{"queries": [ ... ]\}
\end{flushleft}

\vspace{1ex}
\noindent\textbf{User Prompt (input)}
\begin{flushleft}
\ttfamily
Claim: <CLAIM>
\end{flushleft}
\normalsize

\section{Fact-Checking System Evaluation: Annotation Guidelines for NEI claims}
\label{sec:appendixNEI}

\begin{table}[ht]
\centering
\renewcommand{\arraystretch}{1.2}
\begin{tabular}{@{}p{2.65cm}p{4.55cm}@{}} % <-- Fixed width for second column
\toprule
\textbf{Column} & \textbf{Description} \\
\midrule
\texttt{nr} & Row number for easy reference \\
\texttt{claim} & The factual statement to be verified \\
\texttt{true\_label} & Original FEVER dataset label (always “NOT ENOUGH INFO” for these samples) \\
\texttt{predicted\_label} & Our system's prediction (“Supported”, “Refuted”, or “Not Enough Info”) \\
\texttt{found\_evidence} & Evidence found by our system (see format explanations below) \\
\texttt{llm\_explanation} & LLM's reasoning for cases where prediction $\neq$ “Not Enough Info” (should be hidden during annotation) \\
\texttt{human\_annotated} & \textbf{[YOUR TASK]} Mark as “sufficient” or “not sufficient” \\
\texttt{notes} & \textbf{[OPTIONAL]} Space for your reasoning or additional comments \\
\bottomrule
\end{tabular}
\caption{Column structure of our exported CSV file.}
\end{table}

\subsection*{Annotation Instructions}
For each row, you need to evaluate whether the evidence provided is sufficient to support the predicted label.

\subsection*{Step-by-Step Process}
\begin{enumerate}
    \item {Read the claim carefully}
    \begin{itemize}
        \item Understand exactly what factual statement is being made.
    \end{itemize}
    
    \item {Note the predicted label}
    \begin{itemize}
        \item Check if the system predicts \texttt{Supported}, \texttt{Refuted}, or \texttt{Not Enough Info}.
    \end{itemize}
    
    \item {Analyze the found evidence}
    \begin{itemize}
        \item \textbf{For DBpedia evidence:} Assess if the knowledge paths logically support or refute the claim.
        \item \textbf{For Web evidence:} Evaluate the quality and relevance of the snippets, considering source reliability.
    \end{itemize}
    
    \item {Consider additional context (optional)}
    \begin{itemize}
        \item You are welcome to search for additional sources online if needed.
        \item Remember that our system considered many more sources than shown.
    \end{itemize}
    
    \item {Make your judgment}
    \begin{itemize}
        \item In the \texttt{human\_annotated} column, enter:
        \begin{itemize}
            \item \texttt{sufficient} if the evidence adequately supports the predicted label.
            \item \texttt{not sufficient} if the evidence is inadequate, unreliable, or contradictory.
        \end{itemize}
    \end{itemize}
    
    \item {Add notes (optional)}
    \begin{itemize}
        \item Use the \texttt{notes} column to explain your reasoning.
        \item Particularly helpful for borderline cases or when you disagree with the prediction.
    \end{itemize}
\end{enumerate}

\subsection*{Evaluation Criteria}

\paragraph{For \texttt{sufficient} evidence:}
\begin{itemize}
    \item Evidence directly relates to the claim.
    \item Sources appear credible and reliable.
    \item Information is specific and detailed enough to support the conclusion.
    \item Multiple independent sources corroborate the finding (when available).
\end{itemize}

\paragraph{For \texttt{not sufficient} evidence:}
\begin{itemize}
    \item Evidence is tangentially related or off-topic.
    \item Sources appear unreliable or biased.
    \item Information is too vague or general.
    \item Evidence contradicts itself or the predicted label.
\end{itemize}

\begin{figure}[H]
    \centering
    \includegraphics[width=1\linewidth]{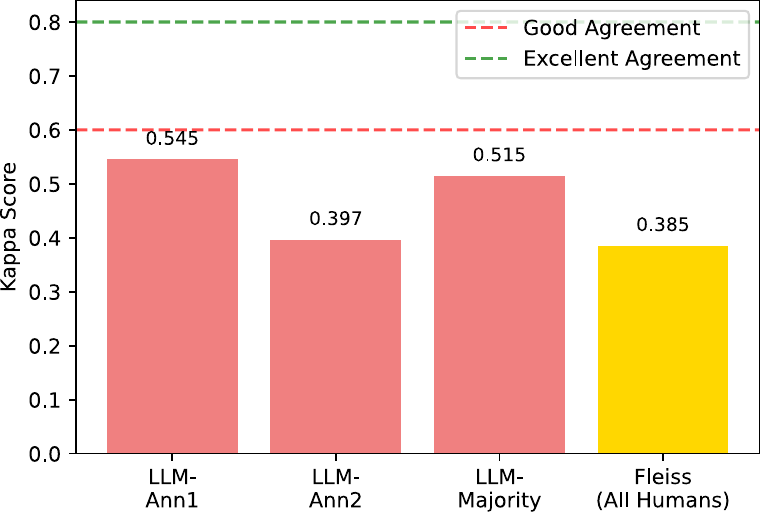}
    \caption{Agreement Scores Comparison. LLM--Human Cohen's $\kappa$ and Human Fleiss' $\kappa$.}
    \label{fig:agreement_score}
\end{figure}

\begin{figure}[H]
    \centering
    \includegraphics[width=1\linewidth]{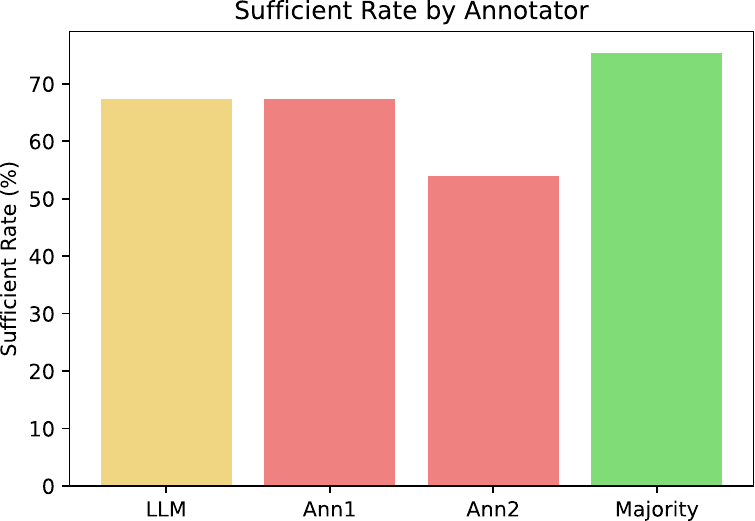}
    \caption{Sufficiency rate differs slightly between annotators.}
    \label{fig:suff_rate}
\end{figure}

\begin{figure}[H]
    \centering
    \includegraphics[width=1\linewidth]{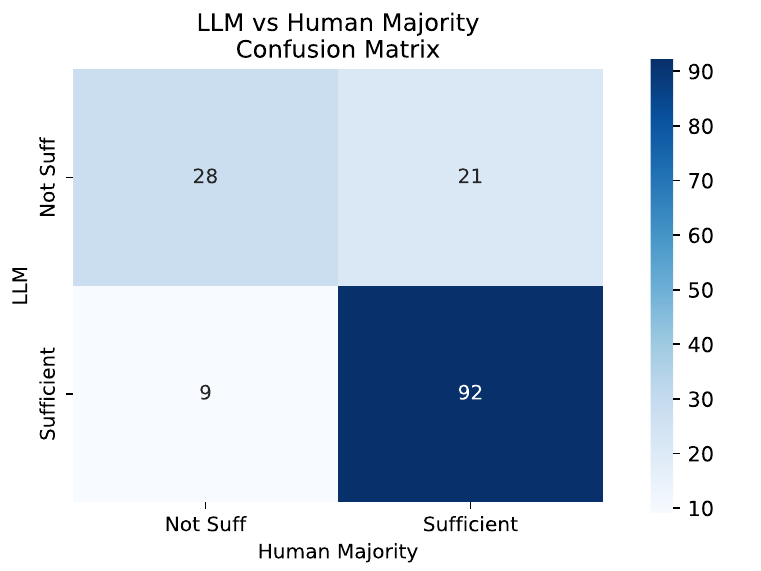}
    \caption{Confusion matrix comparing the LLM's sufficiency judgments with the human majority vote.
}
    \label{fig:confmatrix}
\end{figure}

\end{document}